\documentclass[sigconf]{acmart}

\renewcommand\footnotetextcopyrightpermission[1]{} 
\setcopyright{none}                  

\acmConference[]{}{}{}               
\acmBooktitle{} \acmPrice{} \acmISBN{} \acmDOI{} \acmYear{} \copyrightyear{}

\AtBeginDocument{\providecommand\BibTeX{{\BibTeX}}}

\usepackage{url}
\usepackage{booktabs}
\usepackage{multirow}
\usepackage{graphicx}
\usepackage{amsmath}
\usepackage{array}
\usepackage{placeins}
\usepackage[skip=6.0pt]{caption}  
\AtBeginDocument{%
  \setlength{\textfloatsep}{14pt plus 2pt minus 2pt}
  \setlength{\floatsep}{12pt plus 2pt minus 2pt}
  \setlength{\intextsep}{12pt plus 2pt minus 2pt}
  \setlength{\dbltextfloatsep}{14pt plus 2pt minus 2pt}
  \setlength{\dblfloatsep}{12pt plus 2pt minus 2pt}
}
\usepackage{hyperref}
\usepackage{enumitem}
\title{Beyond ROUGE: N-Gram Subspace Features for LLM Hallucination Detection}

\author{Jerry Li}
\affiliation{\institution{University of California, Riverside} \country{}}
\email{jli793@ucr.edu}

\author{Evangelos Papalexakis}
\affiliation{\institution{University of California, Riverside} \country{}}
\email{epapalex@cs.ucr.edu}

\begin{document}


\begin{abstract}
Large Language Models (LLMs) have demonstrated effectiveness across a wide variety of tasks involving natural language, however, a fundamental problem of hallucinations still plagues these models, limiting their trustworthiness in generating consistent, truthful information. Detecting hallucinations has quickly become an important topic, with various methods such as uncertainty estimation, LLM Judges, retrieval augmented generation (RAG), and consistency checks showing promise. Many of these methods build upon foundational metrics, such as ROUGE, BERTScore, or Perplexity, which often lack the semantic depth necessary to detect hallucinations effectively. In this work, we propose a novel approach inspired by ROUGE that constructs an N-Gram frequency tensor from LLM-generated text. This tensor captures richer semantic structure by encoding co-occurrence patterns, enabling better differentiation between factual and hallucinated content. We demonstrate this by applying tensor decomposition methods to extract singular values from each mode and use these as input features to train a multi-layer perceptron (MLP) binary classifier for hallucinations. Our method is evaluated on the HaluEval dataset and demonstrates significant improvements over traditional baselines, as well as competitive performance against state-of-the-art LLM judges.
\end{abstract}

\maketitle

\pagestyle{plain}     
\fancyhead{}  

\section{Introduction}
Large Language Models (LLMs) have been deployed increasingly across a wide range of applications due to their impressive natural language capabilities. However, a persistent and fundamental challenge that remains across all generative models is hallucinations. These behaviors occur as an LLM must output some sort of answer regardless of the factual truth of the response. Now, with the growing usage of LLMs in important domains such as healthcare, software engineering, customer support, and education, reliably detecting and mitigating these hallucinations has become essential. 

In recent years, there have been various approaches in mitigating LLM hallucinations, such as Chain-of-Thought (CoT) \cite{cothallucinationmitigation}, Retrieval-Augmented Generation (RAG) \cite{ragreducehallucination, rag_knowledge_intensive}, and automatic fact verification systems \cite{factscore, factoolfactualitydetectiongenerative, chain_of_verification}. Combined with scaling, these methods have proved successful in decreasing the hallucination rates of Large Language models. As LLMs hallucinate less, reliably detecting the remaining hallucinations becomes equally important for researchers and users. Lightweight, uncertainty-based detectors use logit-based metrics such as perplexity, entropy, or log probability to flag possible hallucinations and judge the factuality in the generated text  \cite{perplexityOOD, autoregressiveuncertainty, entropy_summarization}. Despite their simplicity and interpretability, they struggle in domain-specific tasks where detecting hallucinations requires fact verification, not confidence estimates. Consistency checks are another growing set of methods where works such as SelfCheckGPT \cite{selfcheck} and INSIDE \cite{insidellmsinternalstates} utilize one LLM as a self-checker or judge for another LLM's responses. However, these types of methods largely depend on the LLM's consistency and can be less interpretable. Retrieval-Augmented Generation (RAG) augments an LLM with external knowledge and has also been used to detect hallucinations via retrieval-based factual validation \cite{sitch_detecting_hallucinations, ragtruth}. This works well on domain-specific tasks but depends on high-quality external sources and retrieval infrastructure. Other alternative recent methods, such as LLM-Check \cite{llmcheck} propose to utilize the internal hidden states and attention maps of these black box models for more accurate real-time analysis. 

Despite the wide range of options for hallucination detection, achieving one that is effective across domains, interpretability, and lightweightness is still difficult. To navigate these trade-offs, we revisit the classic N-Gram based metrics of ROUGE \cite{rouge} and BLEU \cite{bleu}, which were widely used for early summarization and translation tasks \cite{hallucination_survey1}. These often fell short when identifying underlying differences in the semantics of factual and hallucinated content \cite{questeval, faithfulness_summarization}. Newer variants  such as ROUGE-L \cite{mitigate_hallucinations_selfreflection}, BERTScore \cite{bertscore}, BARTScore \cite{yuan2021bartscoreevaluatinggeneratedtext}, and Perplexity \cite{hallucination_survey1} demonstrated improvements but still lag behind more modern approaches like LLM Judges or systems using internal/external sources \cite{hallucination_survey1, hallucination_survey2}. Consequently, pure N-Gram methods have mostly faded from prominence, however, we believe the information source from N-Grams still provides usefulness in the challenge of hallucination detection.

In this work, we propose a method that retains the formation of N-Grams, but changes how this signal of information is used. Instead of an overlap score, we construct an N-dimensional tensor based on the occurrences of N-Grams within LLM-generated documents and apply matrix/tensor decomposition to extract the singular values. These values summarize the latent structure that can distinguish factual and hallucinated generations while maintaining effectiveness and transferability across domains. We demonstrate this by using these singular values as input features for a binary classifier to detect hallucinations. In summary, the contributions of this paper are:

\begin{itemize}
  \item A novel method of extracting latent relationships from N-Gram structured documents through principled matrix and tensor methods.
  \item Experimental results on using singular values from constructed N-Gram Tensors as input features for a hallucination binary classifier for LLM-generated text.
\end{itemize}
\section{Proposed Method: N-Gram Tensor}
\label{sec:method}
We propose representing LLM-generated text using N-Gram tensors, where each entry reflects the count of a specific N-Gram across the text. Notably, we observe that as the reference text length increases, the N-Gram tensor captures more meaningful and discriminative information. This is demonstrated in our experiments with our singular value-based Hallucination classifier.
how
We construct the N-Gram Tensor in three main steps. \textbf{(1)} We first group together texts with the same label, either hallucinated or factual, and convert them into N-Grams. Varying group sizes are used to populate the tensor to be more dense. \textbf{(2)} We then build a vocabulary based on the aggregated text within each group. \textbf{(3)} Finally, we construct the N-Gram tensor, where the Tensor mode corresponds to the N-Gram size and each axis represents the vocabulary. Each entry in the tensor is populated with the frequency of the corresponding N-Gram’s occurrences.

there\subsection{Text Grouping}
Let \( G = \{d_1, d_2, \dots, d_M\} \) denote a grouping of \( M \in \mathbb{N}_{>0} \) text entries.  
For each entry \( d_i \in G \), we extract its sequence of \(N\)-grams, defined as
\[
\mathcal{G}_N(d_i) = \bigl\{ (w_{j}, w_{j+1}, \dots, w_{j+N-1}) \;\big|\; 1 \le j \le |d_i| \bigr\},
\]
where \( w_j \) denotes the \(j\)-the token in \( d_i \), and \( |d_i| \) is the total number of tokens in \( d_i \). We denote the union of all \(N\)-grams across the group as
\[
\mathcal{G}_N(G) = \bigcup_{i=1}^M \mathcal{G}_N(d_i).
\]

\begin{figure*}[!htbp]
  \centering
  \includegraphics[width=\textwidth]{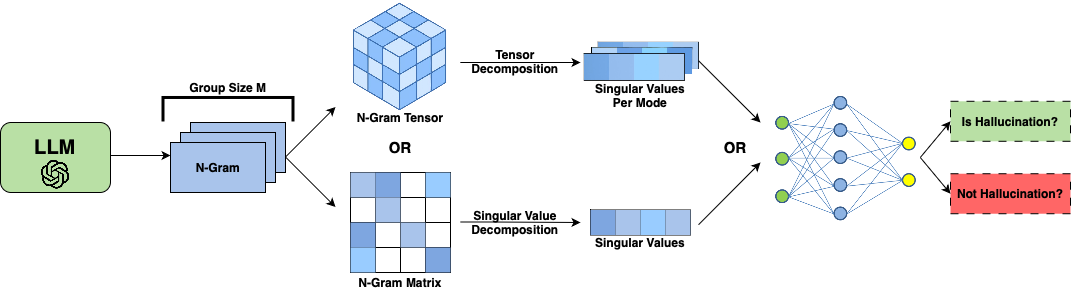}
  \caption{N-Gram Matrix/Tensor creation and input feature extraction process for the binary MLP classifier.}
  \label{fig:ngram_process}
\end{figure*}  

\subsection{Vocabulary Construction}
Using the union of all \(N\)-grams across the group, we can create a group vocabulary $V$, defined as 
\[
V = \{\,v_1,v_2,\dots,v_{|V|}\}
\]
where it's the set of all distinct tokens appearing in any \(N\)-gram in \(\mathcal{G}_N(G)\).

\subsection{N-Gram Tensor Construction}
Using our defined vocabulary, let
\[
\phi:\;V \;\longrightarrow\; \{1,2,\dots,|V|\}
\]
be a bijection that assigns each token \(v\in V\) a unique integer index
\(\phi(v)\).  
Using \(\phi\), every \(N\)-gram
\(g=(w_{1},\dots,w_{N})\) is mapped to its index‑tuple
\[
\pi(g) \;=\;\bigl(\phi(w_{1}),\dots,\phi(w_{N})\bigr)
\;\in\;\{1,\dots,|V|\}^{N}.
\]
We can now define the order‑\(N\) N-Gram frequency tensor
\[
\mathcal{C}\;\in\;\mathbb{N}^{\underbrace{|V|\times\cdots\times|V|}_{N\text{ modes}}},
\]
whose entry at index tuple \( (i_1,\dots,i_N) \) is
\[
\mathcal{C}_{i_1,\dots,i_N}
=
\Bigl|
   \bigl\{\,g\in\mathcal{G}_N(G)\;\bigl|\;
          \pi(g)=(i_1,\dots,i_N)
   \bigr\}
\Bigr|.
\]
In other words, $\mathcal{C}_{i_1,\dots,i_N}$ counts how many times the
$N$-gram $\big(v_{i_1},\allowbreak v_{i_2},\allowbreak \dots,\allowbreak v_{i_N}\big)$
appears in the grouping $G$.

\vspace*{-0.1em}
\begin{table}[h]
\centering
\renewcommand{\arraystretch}{1.0}
\setlength{\tabcolsep}{3pt}
\caption{Comparison of average n-gram counts and response lengths for hallucinated and factual responses across HaluEval subsets.}
\label{tab:halueval_stats}
\begin{tabular}{lcccc}   
\toprule
\textbf{Metric} & \textbf{Dialogue} & \textbf{General} & \textbf{Summary} & \textbf{QA} \\
\midrule
Avg.\ Hallucinated N-Gram & 13.8891 & 64.8307 & 48.3906 & 7.7184 \\
Avg.\ Factual N-Gram      & 8.6047  & 65.0951 & 37.5351 & 1.1782 \\
\midrule
Avg.\ Hallucinated Length & 85.022  & 377.7448 & 325.2583 & 53.669 \\
Avg.\ Factual Length      & 51.751  & 394.2876 & 239.7157 & 13.0485 \\
\bottomrule
\end{tabular}
\end{table}

\section{Experiments}
\subsection{Experimental Setup}

\textbf{Dataset and Evaluation.} In our experiments, we utilize the HaluEval dataset \cite{halueval} along with its three subsets: General, Dialogue, and Summary. These categories provide hallucinated LLM responses and ground truth factual references for user queries across four common task types. The documents in HaluEval are sourced from HotpotQA \cite{hotpotqa}, OpenDialKG \cite{opendialkg}, and CNN/Daily Mail \cite{cnn_dailymail}, ensuring a diverse set of contexts. 

\textbf{Implementation Details.} 
To extract features for our hallucination classifier, we construct N-Gram Tensors following the methodology described in Section~\ref{sec:method}. We then apply tensor decomposition techniques, specifically Singular Value Decomposition (SVD) for matrices and Tucker Decomposition \cite{tucker_parfac} and Canonical Polyadic (CP) Decomposition for higher-order tensors, to extract their singular values using the Tensorly Python library \cite{tensorly}. With Tucker Decomposition, singular values are approximated by flattening and taking the top-k absolute values from the Tucker core tensor, while with CPD, they're extracted through the CP decomposition weights. The resulting singular values across modes are flattened into a fixed-length vector. Each vector is labeled according to whether it corresponds to a hallucinated or non-hallucinated text segment. To ensure uniform input dimensionality, all vectors are cropped or padded to length \(k\), a tunable hyperparameter. This process is applied to each subset of HaluEval, using an 80/20 train–evaluation split within each subset. For experiments done with group sizes larger than 1, we construct the text groups using only each respective train-eval split. Our code is available online.\footnote{\url{https://github.com/Jeli04/NGram-Subspace-Evaluation}}

\subsection{Singular Value Classifier}
We train a multi-layer perceptron (MLP) for binary classification using the extracted singular value vectors as input. Training is done separately across four different text group sizes of 1, 5, 20, and 40 for the N-Gram Tensor. The MLP consists of four fully connected layers with hidden dimensions of 48, 64, 32, and 1. ReLU activation is applied between layers, and the model is optimized using binary cross-entropy (BCE) loss. For each experiment across all HaluEval subsets, we train for 20 epochs with a learning rate of $1e-4$.

\begin{figure}[!htbp]
  \centering
  \includegraphics[width=0.45\textwidth]{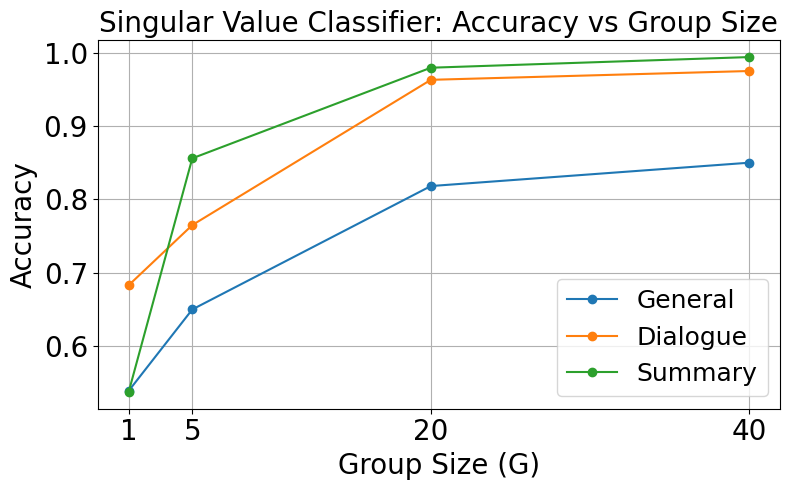}
  \caption{Performance comparison as text group size increases in N-Gram Matrix construction. Larger group sizes consistently improve binary classifier performance across all HaluEval subsets.}
  \label{fig:svd_groups}
\end{figure}

\subsection{Results}
In this section, we present the evaluation results of our MLP binary classifier across various text group sizes and compare it to baseline metrics. The experiments were conducted across three subsets of HaluEval: General, Dialogue, and Summary, with varying text group sizes of 1, 5, 20, and 40. The group sizes indicate the number of text samples aggregated into a single matrix or tensor during the creation of our N-Gram Tensor. For group sizes 1 and 5, a feature size (hyperparameter k) of 20 was used, while for group sizes 20 and 40, a feature size of 40 was chosen. These feature sizes were determined through trial and error in experimentation. The QA subset was left out due to a noticeable, strong textual bias where the factual answers were consistently shorter phrases.

\subsubsection{Baselines}
Table~\ref{svd-results} summarizes our experimental results. We first evaluated three baseline metrics: Perplexity, Rouge-L, and BERTScore. We first calibrate thresholds on the train subsets with Youden’s J Statistic, a measure maximizing the difference between true positive and false positive rates. Afterwards, we apply a simple binary classification using this calibrated threshold on the evaluation subset. Between these three, Perplexity consistently achieved the highest scores across all HaluEval subsets. However, it still fell short compared to our method when larger group sizes were employed. Additionally, we also use the HaluEval code repository to evaluate a GPT-4o \cite{gpt4o} LLM judge. 

\subsubsection{Singular Value Decomposition Features} 
Table~\ref{svd-results} demonstrates the results when we train the binary MLP classifier on SVD singular values created from 2-Gram Matrices. The results consistently improve as the group size increases, highlighting an underlying trend of denser and more populated matrices yielding better performances as shown in Figure~\ref{fig:svd_groups}. For group sizes of 1, our method performed worse than Perplexity in the General and Summary subsets, however, starting from a group size of 5, our method significantly surpassed Perplexity and eventually even outperformed the GPT-4o based LLM Judge, notably at larger group sizes of 20 and 40. This supports that aggregating more context into singular value feature representations effectively improves the classifier's performance.

\begin{figure}[!htbp]
  \centering
  \includegraphics[width=0.5\textwidth]{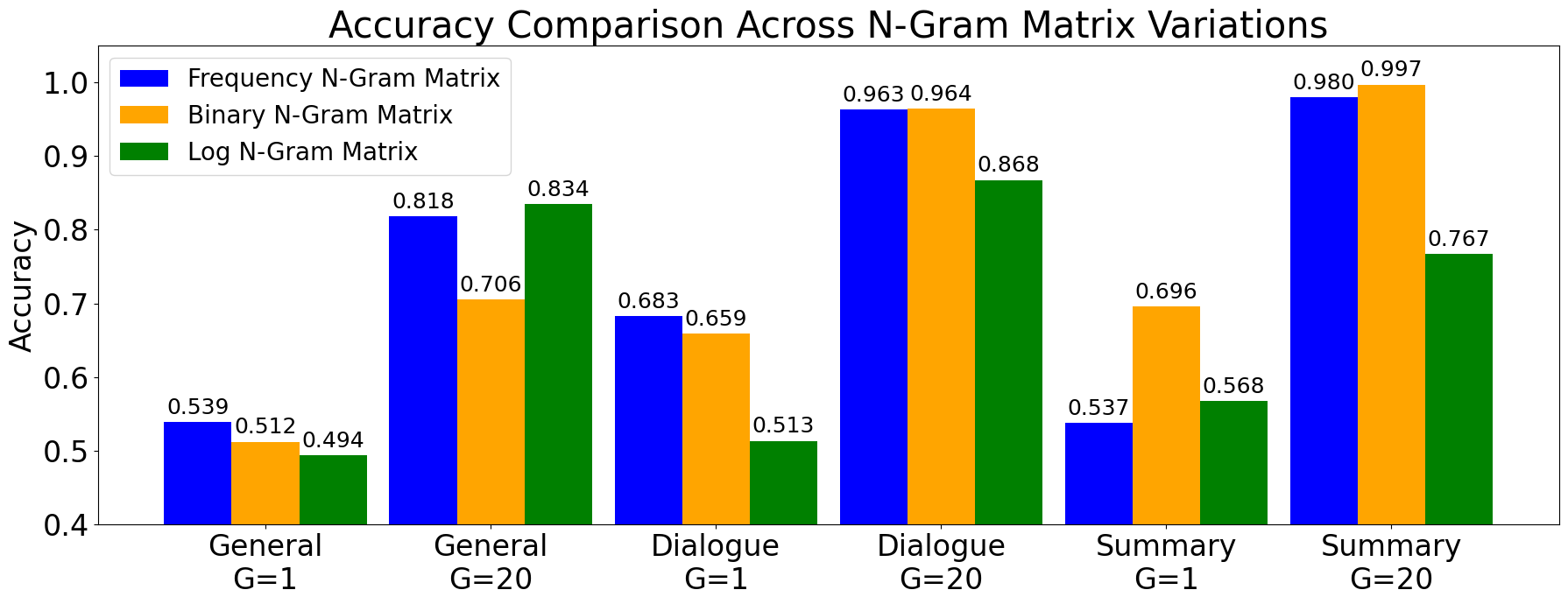}
  \caption{Comparison of performances between Binary, Log Frequency, and Frequency N-Gram matrices. Alternative methods demonstrate the robustness of the Frequency matrix against textual biases across dataset categories.}
  \label{fig:comparisons}
\end{figure}

\begin{table}[!htbp]
\centering
\renewcommand{\arraystretch}{1.1}
\setlength{\tabcolsep}{4pt}
\begin{tabular}{l|lcccc}
\toprule
\textbf{Dataset} & \textbf{Method} & \textbf{AUROC} & \textbf{AUPR} & \textbf{F1} & \textbf{Accuracy} \\
\midrule

\multirow{8}{*}{General}
  & Perplexity   & 0.5113 & 0.5054 & 0.5523 & 0.5552 \\
  & Rouge-L      & 0.4784 & 0.4831 & 0.5015 & 0.5153 \\
  & BERTScore    & 0.5148 & 0.5186 & 0.4907 & 0.4877 \\
  & GPT4o Judge    & --     & --     & 0.3233     & 0.7910     \\
  \cmidrule{2-6}
  & SVD-G1       & 0.5399 & 0.5213 & 0.5810 & 0.5390 \\
  & SVD-G5       & 0.6500 & 0.6041 & 0.6051 & 0.6500 \\
  & SVD-G20      & 0.8182 & 0.7475 & 0.8305 & 0.8182 \\
  & SVD-G40      & --     & --     & 0.8619 & 0.8500 \\
\midrule

\multirow{8}{*}{Dialogue}
  & Perplexity   & 0.5000 & 0.4996 & 0.5525 & 0.6275 \\
  & Rouge-L      & 0.4947 & 0.4921 & 0.5097 & 0.5045 \\
  & BERTScore    & 0.4762 & 0.4797 & 0.5197 & 0.5275 \\
  & GPT4o Judge    & --     & --     & 0.7617     & 0.7031     \\
  \cmidrule{2-6}
  & SVD-G1       & 0.6830 & 0.6213 & 0.7015 & 0.6830 \\
  & SVD-G5       & 0.7650 & 0.7112 & 0.7516 & 0.7650 \\
  & SVD-G20      & 0.9630 & 0.9400 & 0.9635 & 0.9630 \\
  & SVD-G40      & --     & --     & 1.0000 & 0.9750 \\
\midrule

\multirow{8}{*}{Summary}
  & Perplexity   & 0.4990 & 0.4990 & 0.5564 & 0.6000 \\
  & Rouge-L      & 0.4863 & 0.4875 & 0.5003 & 0.4995 \\
  & BERTScore    & 0.5012 & 0.5015 & 0.4972 & 0.4955 \\
  & GPT4o Judge    & --     & --     & 0.6874     & 0.7480     \\
  \cmidrule{2-6}
  & SVD-G1       & 0.5375 & 0.5197 & 0.6247 & 0.5375 \\
  & SVD-G5       & 0.8560 & 0.8018 & 0.8577 & 0.8560 \\
  & SVD-G20      & 0.9795 & 0.9775 & 0.9792 & 0.9795 \\
  & SVD-G40      & 0.9940 & 0.9935 & 0.9940 & 0.9940 \\
\bottomrule
\end{tabular}
\caption{%
    Performance comparison between baseline methods and our SVD-feature-based classifier across HaluEval subsets. SVD-GX indicates the group size X used to construct the matrix. Empty entries are due to constraints with memory when constructing tensors.
}
\label{svd-results}
\end{table}

\begin{table}[h]
\centering
\renewcommand{\arraystretch}{1.1}
\setlength{\tabcolsep}{4pt}
\begin{tabular}{l|lcccc}
\toprule
\textbf{Dataset} & \textbf{Method} & \textbf{AUROC} & \textbf{AUPR} & \textbf{F1} & \textbf{Accuracy} \\
\midrule

\multirow{8}{*}{General}
  & TUCKER-G1    & 0.5530 & 0.5285 & 0.6263 & 0.553  \\
  & TUCKER-G5    & 0.6165 & 0.5714 & 0.6224 & 0.6165 \\
  & TUCKER-G20   & 0.6510 & 0.5978 & 0.6551 & 0.651  \\
  & TUCKER-G40   & --     & --     & --     & --     \\
  \cmidrule{2-6}
  & CPD-G1       & 0.6285 & 0.5802 & 0.6352 & 0.6285 \\
  & CPD-G5       & 0.7025 & 0.6488 & 0.6806 & 0.7025 \\
  & CPD-G20      & --     & --     & --     & --     \\
  & CPD-G40      & --     & --     & --     & --     \\
\midrule

\multirow{8}{*}{Dialogue}
  & TUCKER-G1    & 0.6735 & 0.6148 & 0.6853 & 0.6735 \\
  & TUCKER-G5    & 0.7840 & 0.7646 & 0.7394 & 0.784  \\
  & TUCKER-G20   & 0.8310 & 0.7638 & 0.8401 & 0.831  \\
  & TUCKER-G40   & 0.9010 & 0.8670 & 0.8993 & 0.901  \\
  \cmidrule{2-6}
  & CPD-G1       & 0.6045 & 0.5626 & 0.6143 & 0.6045 \\
  & CPD-G5       & 0.6540 & 0.6018 & 0.6462 & 0.6540 \\
  & CPD-G20      & 0.6455 & 0.5922 & 0.6606 & 0.6455 \\
  & CPD-G40      & 0.7130 & 0.6417 & 0.7491 & 0.7130 \\
\midrule

\multirow{8}{*}{Summary}
  & TUCKER-G1    & 0.5530 & 0.5285 & 0.6263 & 0.553  \\
  & TUCKER-G5    & 0.6165 & 0.5714 & 0.6224 & 0.6165 \\
  & TUCKER-G20   & 0.6510 & 0.5978 & 0.6551 & 0.651  \\
  & TUCKER-G40   & --     & --     & --     & --     \\
  \cmidrule{2-6}
  & CPD-G1       & 0.6285 & 0.5802 & 0.6352 & 0.6285 \\
  & CPD-G5       & 0.7025 & 0.6488 & 0.6806 & 0.7025 \\
  & CPD-G20      & --     & --     & --     & --     \\
  & CPD-G40      & --     & --     & --     & --     \\
\bottomrule
\end{tabular}
\caption{Comparison of TUCKER-based and CPD-based methods across HaluEval subsets. Empty entries are due to constraints with memory when constructing tensors.}
\label{tab:tucker_cpd_results}
\end{table}

\subsubsection{Performance from Textual Bias}
Our evaluation across the various subsets of HaluEval demonstrates significant performance improvements as the group size increases. Specifically, for larger group sizes (20 and 40), we observed exceptionally high results, raising concerns about the potential influence of unwanted textual biases within each subset. Given that each subset naturally contains distinct characteristics and potential biases, as seen with N-Gram lengths in Table~\ref{tab:halueval_stats}, it is important to assess whether our binary classifier is leveraging these biases unintentionally.

To investigate this, we created two additional variations of our N-Gram Matrix. Instead of tracking N-Gram frequencies directly, these alternative matrices utilized binary and log-frequency representations. The binary based matrix eliminates frequencies from being measured completely, while the log-frequency normalizes the frequencies, reducing the impact of outliers. These approaches aim to minimize reliance on potentially biased N-Grams specific to each subset category. As shown in Figure 3, we compare these N-Gram matrix variations across group sizes 1 and 20 for all three subsets.

Interestingly, certain group and subset combinations demonstrated comparable or even improved performance with the alternative representations. These experiments suggest that our method using frequency based N-Gram matrices is robust enough and minimally influenced by category-specific textual biases. 

\section{Conclusion}
In this work, we proposed a novel approach for capturing latent structure from N-Gram-based matrices and tensors to distinguish hallucinated from factual LLM-generated text. By using singular values from these representations as input to a binary MLP classifier, our method matches or outperforms classical metrics and a modern LLM judge across multiple settings. To confirm robustness and rule out textual bias, we evaluated alternative representations, demonstrating the reliability of our approach.

Although hallucination detection has made significant advances, there remains substantial room for exploring the underlying mechanisms of hallucination. Future work should focus on improving the interpretability of our N-Gram Matrices and Tensors to better understand how they capture differences between hallucinated and factual text. Additionally, investigating how hallucinations differ across text categories could offer valuable insights into the nature of LLM hallucinations.

\section{Generative AI Disclosure}
Other than latex formatting, this document was created without assistance from AI tools. The content has been reviewed and edited by a human. Please contact the author if you have concerns or comments.

\section{Acknowledgements}
This research was supported in part by the National Science Foundation under CAREER grant no. IIS 2046086, grant no. No. 2431569 and CREST Center for Multidisciplinary Research Excellence in CyberPhysical Infrastructure Systems (MECIS) grant no. 2112650, and by the University Transportation Center for Railway Safety (UTCRS) at UTRGV through the USDOT UTC Program under Grant No. 69A3552348340. The views and conclusions contained in this document are those of the authors and should not be interpreted as representing the official policies, either expressed or implied, of the U.S. Government.

\bibliographystyle{ACM-Reference-Format}
\bibliography{references}

\end{document}